# A new filter for dimensionality reduction and classification of hyperspectral images using GLCM features and mutual information

## Hasna Nhaila*, Elkebir Sarhrouni and Ahmed Hammouch

Electrical Engineering Department,
LRGE, ENSET,
Mohammed V University in Rabat, 10100, Morocco
Email: hasnaa.nhaila@gmail.com
Email: sarhrouni436@yahoo.fr
Email: hammouch_a@yahoo.com
*Corresponding author

**Abstract:** Dimensionality reduction is an important preprocessing step of the hyperspectral images classification (HSI), it is inevitable task. Some methods use feature selection or extraction algorithms based on spectral and spatial information. In this paper, we introduce a new methodology for dimensionality reduction and classification of HSI taking into account both spectral and spatial information based on mutual information. We characterise the spatial information by the texture features extracted from the grey level cooccurrence matrix (GLCM); we use Homogeneity, Contrast, Correlation and Energy. For classification, we use support vector machine (SVM). The experiments are performed on three well-known hyperspectral benchmark datasets. The proposed algorithm is compared with the state of the art methods. The obtained results of this fusion show that our method outperforms the other approaches by increasing the classification accuracy in a good timing. This method may be improved for more performance.

**Reference** to this paper should be made as follows: Nhaila, H., Sarhrouni, E. and Hammouch, A. (2018) 'A new filter for dimensionality reduction and classification of hyperspectral images using GLCM features and mutual information', *Int. J. Signal and Imaging Systems Engineering*, Vol. 11, No. 4, pp.193–205.

**Biographical notes:** Hasna Nhaila received her Master degree in Electrical Engineering from ENSET, Rabat Mohammed V University, Morocco, in 2012. She is a research student of Sciences and Technologies of the Engineer in ENSIAS, Research Laboratory of Electrical Engineering LRGE, Research Team in Computer and Telecommunication at ENSET, Mohammed V University, Rabat, Morocco. Her interests, in the context of national doctoral thesis, are in hypespectral images classification, pattern recognition and dimensionality reduction.

Elkebir Sarhrouni graduated in Electronic and Industrial Computer Aggregation in 1995. Since 2003, he is a Member of the laboratory LRIT (Unit associated with the CNRST, FSR, Mohammed V University, Rabat, Morocco). He acquired his PhD in Computer and Telecommunication from Mohammed V-Agdal University, Rabat, Morocco in 2014. His domains of interest include signal processing and embedded systems.

Ahmed Hammouch received his Master degree and the PhD in Automatic, Electrical, Electronic by the Haute Alsace University of Mulhouse (France) in 1993 and the PhD in Signal and Image Processing by the Mohammed V University of Rabat in 2004. From 1993 to 2013, he was a professor in the Mohammed V University in Morocco. Since 2009 he manages the Research Laboratory in Electronic Engineering. He is an author of several papers in international journals and conferences. His domains of interest include multimedia data processing and telecommunications. He is currently Head of department for Scientific and Technical Affairs in the National Center for Scientific and Technical Research in Rabat.

# A New filter for dimensionality reduction and classification of Hyperspectral images using GLCM features and mutual information

*Abstract*—**Dimensionality reduction is an important issue before the classification of the hyperspectral images (HSI), it's inevitable task. Some methods use feature selection or extraction algorithms based on spectral and spatial information. But in this Paper, we introduce a new methodology for dimensionality reduction and classification of Hyperspectral Images taking into account both spectral and spatial information based on mutual information. We extract texture features by Gray Level Co-occurrence Matrix (GLCM), we use Homogeneity, Contrast, Correlation and Energy. For classification, we use Support vector machine (SVM). The experiments are performed on AVIRIS HSI 92AV3C to validate the proposed method. The obtained results of this fusion are better than spectral classification. This method may be improved for more performance.**

## I. INTRODUCTION

THE Hyperspectral images consist to acquire spectra for all image pixels, they provide more than a hundred of bands of the same region with more detailed information , the rich availability of hyperspectral data increases the discrimination of spectral signatures compared to multispectral images. Thus, it has been used as an important mean for Earth observation and exploration and other applications (Nhaila, Sarhrouni and Hammouch 2014). However, this large amount of data causes difficulties of storage, transmission and possesses new challenges in the processing systems due to the curse of dimensionality. For this purpose, the dimensionality reduction of the hyperspectral images becomes a priority preprocessing of classification. Several works have been developed in this area to explain the need to eliminate the redundant or irrelevant bands by features selection or extraction (Guo et al. 2006).

In this study, we propose a new algorithm to combine spectral information with the texture features extracted from GLCM to improve the results of HSI classification using the mutual information. It's an improved method of the process proposed in (Sarhrouni, Hammouch and Aboutajdine 2012) where just spectral information of the images was used based on mutual information.

The rest of this paper is organized as follows. The second section provides an overview of texture based classification methods and the related works. In section 3, we outline the proposed methodology using both mutual information and texture features. The data set and experimental results are presented and discussed in section 4. Finally section 5 concludes our work.

## II. OVERVIEW OF TEXTURE BASED CLASSIFICATION METHODS

Texture is an important characteristic for the images classification. Thus, many approaches have been applied for texture analysis according to the processing algorithms and can be classified in three categories namely, Spectral, Structural and Statistical methods (Gonzalez and Woods 2002) as shown in Figure 1.

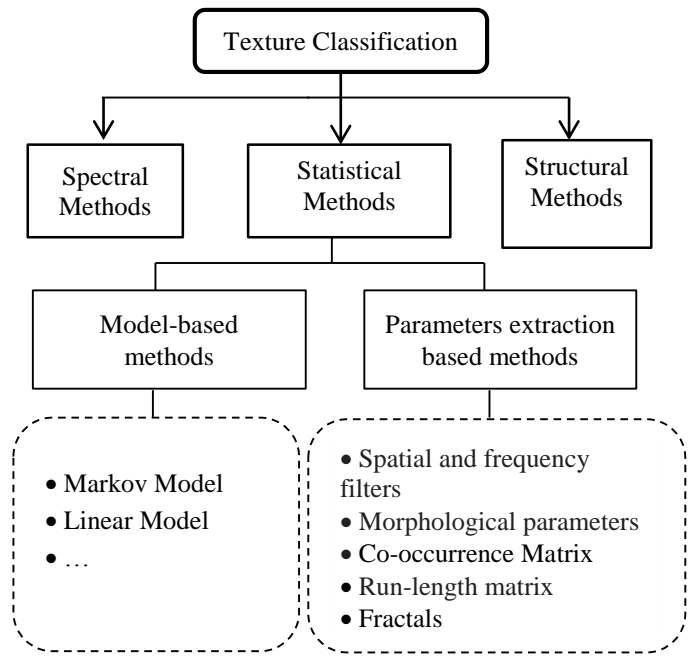

Fig. 1.Texture Classification Methods

Spectral methods consist to convert an image from spatial to frequency domain and vice-versa using filter responses such as filter banks. For this, several works have been successfully developed (Malik et al. 1999), (Varma and Zisserman 2002). Studies on Structural methods are still limited due to their complexity than statistical ones especially when the textures are weakly structured (Benediktsson et al. 2005). Statistical methods on the other hand, analyze the spatial distribution of gray values based on statistical proprieties of images, there are two categories: the model based methods for example Markov model (Li, Bioucas-Dias and Plaza 2012), (Yongqing and Yingling 2006) and features extraction such as Fractals (Pentland 1984), (Unser 1986).

Harralick (1979) proposed the extraction of second order statistics from images using GLCM method (Arivazhagan and Ganesan 2006), it's one of the most useful of statistical methods (Ohanian and Dubes 1992). Tsai and Lai (2013) extended GLCM to third order of texture measures. Shi and Healy (2003) used Gabor filters on bands with reduced dimensionality to compute texture features. In this work, for texture analysis, we used Co-occurrence matrix to extract texture features.

## III. PROPOSED ALGORITHM

### A. General principle

For HSI classification, Sarhrouni et al. (2012) proposed a filter approach using mutual information "Algorithm1". In this work, we reproduce this approach and we propose a new filter that combines spectral and spatial information where we use four texture features namely energy, homogeneity, contrast and correlation extracted from GLCM matrix to improve the classification results "Algorithm2". We applied the mutual information to select the optimal bands prior to the classification; this pre-treatment allows to reducing the dimensionality of the image by eliminating the irrelevant and redundant attributes. The classifier used in this paper is SVM. To approve the effectiveness of our method, we'll use the HSI AVIRIS 92AV3C.

The flow chart of the proposed methodology is illustrated in Fig. 2. The detailed process of this method is described in this section.

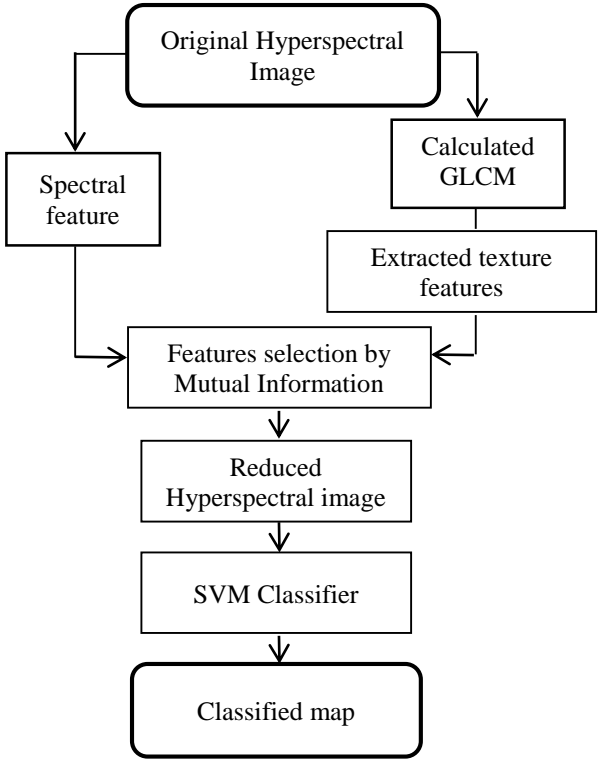

Fig. 2. Flow chart of the proposed methodology

### B. Retained method for texture features extraction

The majority of classification methods use the spectral dimension where each pixel is considered as vector of attributes and may be used directly as an input of the classifier. In our research, we exploit the spatial relationship of pixels using the Co-occurrence matrix method GLCM. It's considered as the reference of images classification since it was proposed by Haralick, Shanmugam and Dinstein (1973).

The size of this matrix is equal to the number of gray levels in the image; the distribution depends on the distance d between two pixels in four directions $\theta$: 0°, 45°, 90° and 135°. The following figure 3, shows an example of calculation of the co-occurrence matrix from 5 x 5 image composed of 3 gray levels (0,1,2) in the case of $d = (0,1)$.

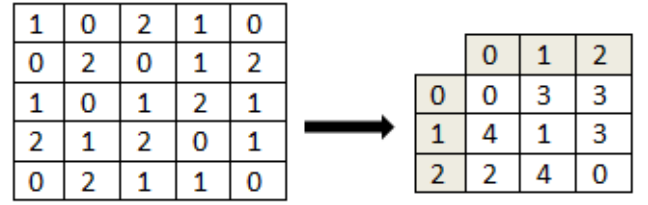

Fig. 3. Image 5x5 with 3 gray levels and the corresponding co-occurrence matrix

From the GLCM created, various features can be extracted, in our case, we used the following four:

- **Contrast**

It measures the intensity contrast between two pixels. For a constant image, contrast is 0.
The contrast is calculated through the equation 1.

$$Contrast = \sum_i \sum_j (i-j)^2 P(i,j) \qquad (1)$$

- **Correlation**

Correlation measures the gray level linear dependence between pixels. It's NaN for a constant image.

$$Correlation = \sum_i \sum_j \frac{(i-\mu_i)(j-\mu_j)}{\sigma_i \sigma_j} P(i,j) \qquad (2)$$

- **Energy**

The energy E measures the sum of squared elements in the GLCM using the equation 3, E=1 if the image is constant.

$$Energy = \sum_i \sum_j P(i,j)^2 \qquad (3)$$

- **Homogeneity**

The homogeneity measures the closeness of the distribution of elements in the GLCM diagonal through the following equation 4. It has maximum value when all elements of the image are same.

$$Homogeneity = \sum_i \sum_j \frac{1}{1+(i-j)^2} P(i,j) \qquad (4)$$

Where:

$P(i,j)$: Element i,j of the GLCM

μ: The mean of the GLCM

σ: The standard deviation.

### C. Retained method for band selection: Mutual Information

The mutual information MI is a statistical measure of the similarity between two random variables: a reference (in our case the ground truth map) that we note A and each band noted B.
The MI between A and B is given as:

$$I(A,B) = \sum log_2 p(A,B) \frac{p(A,B)}{p(A).p(B)} \qquad (5)$$

In relation with Shanon entropy, the MI can be equivalently expressed as:

$$I(A, B) = H(A) + H(B) - H(A, B) \qquad (6)$$

This expression is illustrated in the following Venn diagram, see figure 4.

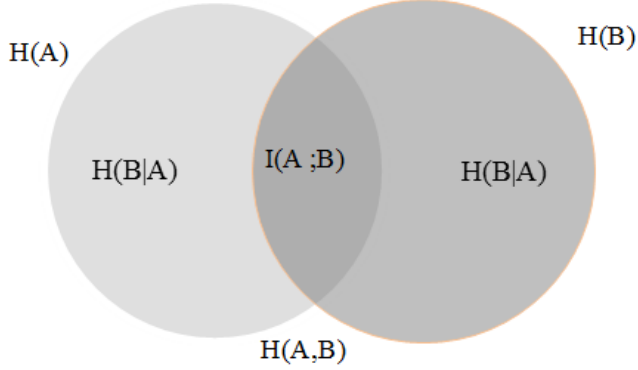

*Fig. 4.  Venn diagram*

For our experiments, high value of MI means a large similarity between the ground truth map and the band, where low MI indicates a small similarity and zero MI shows that the two variables A and B are independent.

### D.  The reproduced algorithm

The algorithm of HSI dimensionality reduction and classification proposed by Sarhrouni et al. (2012) used the mutual information, its selection "Algorithm1" is described as follows.

**Algorithm 1:** Let SS be the ensemble of bands already selected and S the band candidate to be selected. SS is initially empty; R the ensemble of band candidate, it contains initially all bands (1..220). MI is initialized with a value MI*, X the number of bands to be selected and Th the threshold controlling redundancy:

1) Select the first band : $S = argmax_s \, MI(s)$;
$SS \leftarrow S$;
$C_{est0} = Band(S)$;
**while** $|SS| < X$ **do**
Select $band\ index_s\ S = argmax_s \, MI(s)$ and $R \leftarrow R \backslash S$;

$$C_{est} = \frac{C_{est0} + Band(S)}{2};$$

$C_{est} = Build\_estmated\_C$ ;
$MI = Mutual\_Information(Gt, C_{est})$
**if** $MI > MI^* + Threshold$ **then**
$MI^* = MI$;
$C_{est0} = C_{est}$;
$SS \leftarrow SS \cup S$;
**end if**
**end while**

### E.  The new improved Algorithm

In this method, we will use four spatial features that characterize the texture extracted via the Grey Level Co-occurrence Matrix (GLCM) namely energy, contrast, homogeneity and correlation then we will combine all these characteristics in the same process to improve the classification results. We performed our experiments with SVM classifier (Unser 1986).

So our proposed selection "Algorithm2" is as follows:

**Algorithm 2:**  Let GLCM be the matrix containing the four texture features of the HSI: H the homogeneity, C the contrast, Cor the correlation and E the energy; SS the ensemble of bands already selected and S the band candidate. SS is initially empty; R the ensemble of bands candidate, it contains initially all bands (1..220). MI is initialized with a value MI*, X the number of retained bands and Th the threshold controlling redundancy:

1) Features Extraction
Calculate the GLCM of bands and the GT.
$C \leftarrow GLCM(1); Cor \leftarrow GLCM(2); E \leftarrow GLCM(3);$
$H \leftarrow GLCM(4);$
2) Select the first band to initialize C_est:
Select $band\ index_s\ S = C(s) \, or \, Cor(s) \, or \, E(s) \, or \, H(s);$
$SS \leftarrow S;$
$R \leftarrow R \backslash S;$
$C_{est0} = Band(S);$
3) Selection process:
**while** $|SS| < X$ **do**
Select $band\ index_s\ S = H(s) \, or \, C(s) \, or \, Cor(s) \, or \, E(s)$
$and \, R \leftarrow R \backslash S;$

$$C_{est} = \frac{C_{est0} + Band(S)}{2};$$

$C_{est} = Estmated\_C$ ;
$MI = Mutual\_Information(Gt, C_{est})$
**if** $MI > MI^* + Threshold$ **then**
$MI^* = MI$;
$C_{est0} = C_{est}$;
$SS \leftarrow SS \cup S$;
**end if**
**end while**

## IV.  Experimental Results

### A.  The Data Set used

To test the efficiency of the algorithms aforesaid, we have chosen to apply them on the Hyperspectral image AVIRIS 92AV3C obtained from the Airborne Visible Infrared Imaging Spectrometer for the scene Indiana pines in the north Indiana in 1992, it contains 220 bands with 145x145 pixels which are labeled on 16 classes as shown in figure 5.

Two-thirds of this image are covered by agricultural land and the one-third by forest or other built structures. 50% of labeled pixels are selected to be used in training and the other 50% will be applied for the classification testing.

The classifier used is SVM and the implementation of the programs was made using the scientific programming language "Matlab".

| | The accuracy (%) of classification for numerous thresholds | | | | |
|---|---|---|---|---|---|
| | -0,0200 | -0,0100 | -0,0050 | -0,0040 | 0,0000 |
| **2** | 53,61 | 53,61 | 53,61 | 53,61 | 53,61 |
| **3** | 54,37 | 54,37 | 54,37 | 54,37 | 54,37 |
| **4** | 54,8 | 54,8 | 54,8 | 54,8 | |
| **12** | 64,56 | 64,56 | 63,9 | 63,9 | |
| **14** | 64,76 | 64,76 | 65,44 | 64,93 | |
| **18** | 66,71 | 66,71 | 67,47 | 67,86 | |
| **20** | 68,09 | **68,09** | 68,54 | 68,38 | |
| **25** | 74 | 74,05 | 78,28 | **78,39** | |
| **35** | 78,24 | 78,63 | 80,77 | | |
| **36** | 78,01 | 79,17 | 81,39 | | |
| **40** | 79,19 | 81,41 | | | |
| **45** | 81,9 | 82,12 | | | |
| **50** | **82,37** | 82,86 | | | |
| **53** | 82,84 | 83,46 | | | |
| **60** | 83,99 | 84,32 | | | |
| **70** | 87,28 | | | | |
| **75** | 87,08 | | | | |
| **80** | 86,56 | | | | |

(left margin label: **Number of retained Bands**)

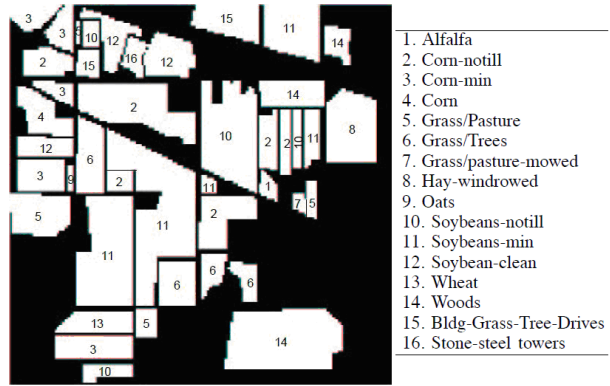

Fig. 5. The Ground Truth map of AVIRIS 92AV3C

Legend:
1. Alfalfa
2. Corn-notill
3. Corn-min
4. Corn
5. Grass/Pasture
6. Grass/Trees
7. Grass/pasture-mowed
8. Hay-windrowed
9. Oats
10. Soybeans-notill
11. Soybeans-min
12. Soybean-clean
13. Wheat
14. Woods
15. Bldg-Grass-Tree-Drives
16. Stone-steel towers

## B. Results

The table I gives the reproduced results of classification Tx for different thresholds Th using spectral features by mutual information "Algorithm1" proposed by Sarhrouni et al. (2012).

Table II represents the accuracy classification of the proposed method "Algorithm 2" based on the combination of spectral information and the spatial features.

TABLE I
REPRODUCED RESULTS FOR THE "ALGORITHM1"

| | The accuracy (%) of classification for numerous thresholds | | | | |
|---|---|---|---|---|---|
| | -0,0200 | -0,0100 | -0,0050 | -0,0040 | 0,0000 |
| **2** | 47,44 | 47,44 | 47,44 | 47,44 | 47,44 |
| **3** | 47,87 | 47,87 | 47,87 | 47,87 | 48,92 |
| **4** | 49,31 | 49,31 | 49,31 | 49,31 | |
| **12** | 56,30 | 56,30 | 56,30 | 56,30 | |
| **14** | 57,00 | 57,00 | 57,00 | 57,00 | |
| **18** | 59,09 | 59,09 | 59,09 | 62,61 | |
| **20** | 63,08 | **63,08** | 63,08 | 63,55 | |
| **25** | 66,12 | 64,89 | 64,89 | **65,38** | |
| **35** | 76,06 | 74,72 | 75,59 | | |
| **36** | 76,49 | 76,60 | 76,19 | | |
| **40** | 78,96 | 79,29 | | | |
| **45** | 80,85 | 81,01 | | | |
| **50** | **81,63** | 81,12 | | | |
| **53** | 82,27 | 86,03 | | | |
| **60** | 82,74 | 85,08 | | | |
| **70** | 86,95 | | | | |
| **75** | 86,81 | | | | |
| **80** | 87,28 | | | | |

(left margin label: **Number of retained Bands**)

## C. Discussion

From the above tables I and II it's seen that:

The use of both spectral spatial features proposed in the new method "Algorithm2" gives better classification results compared with the use of only spectral information proposed in "Algorithm1", for example in the case of Th = -0.01 with X=20bands, the classification accuracy of the first algorithm is 63.08% (table I) and it achieves 68.09% for the proposed algorithm2 as shown in the table II.

The threshold affects widely the classification results:

• First, for high values, in the range of (-0.004 to 0), few bands are retained because we don't allow the redundancy. The highest accuracies are obtained using the proposed algorithm, for example in the case of Th=-0.004 with just 25 retained bands, the new proposed "Algorithm3" gives 78.39% which is better than "Algorithm1" by 13.01%.

• Second, for medium values, in the range of (-0.005 to -0.02) where we permit some redundancy, also the proposed method produced the best reslts. For example, in the case of Th=-0.02 with 50 retained bands, it achieves 82.37% for "Algorithm2", where "Algorithm1" gives 81.63%.

• Third, for more redundancy, for Th less than (-0.02), we didn't have interesting results for more than 80 bands.

Effectively, Table II shows the effectiveness selection using the four texture features extracted from the GLCM namely contrast, correlation, homogeneity and energy proposed in this work "Algorithm2" compared with Algorithm1 based only on spectral features.

The Graph in the following figure6 also illustrates this positive effect of the proposed algorithm, it shows the classification results obtained for different number of retained bands X in the case of Th=-0.02.

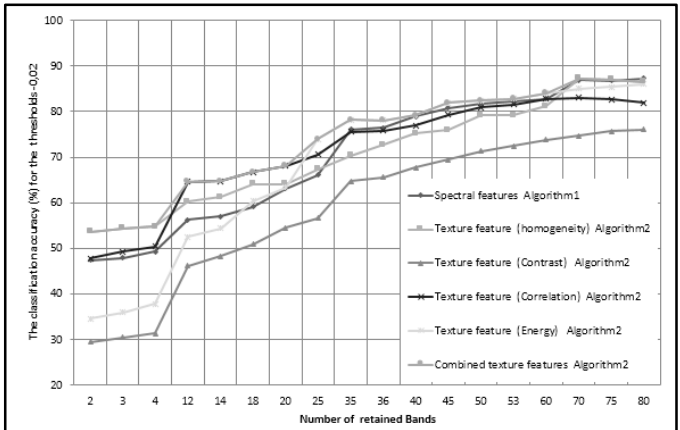

Fig. 6. Classification accuracy of the three Algorithms for Th=-0.02

All the previous results were presented considering all the 16 classes of the AVIRIS Indiana Pine data set. Now we represent in the following table III, the classification accuracy of each class for Th=-0.02 obtained using the proposed algorithm.

TABLE III
THE CLASSIFICATION ACCURACY OF EACH CLASS OBTAINED USING THE PROPOSED ALGORITHM FOR Th=-0.02

| Class | Total pixels | Classification accuracy (%) for Th= -0.02 | | | |
|---|---|---|---|---|---|
| | | Contrast X=80 | Correlation X=70 | Energy X=80 | Homogeneity X=70 |
| 1 | 54 | 39,13 | 78,26 | **86,96** | 82,61 |
| 2 | 1434 | 68,48 | 74,06 | **81,87** | 79,64 |
| 3 | 834 | 63,79 | 80,58 | 80,10 | **82,97** |
| 4 | 234 | 27,35 | 58,97 | **72,65** | 69,23 |
| 5 | 497 | 76,83 | 89,02 | **92,68** | 90,65 |
| 6 | 747 | 91,62 | 94,41 | **96.37** | 95,81 |
| 7 | 26 | 38,46 | 61,54 | 76,92 | **84,62** |
| 8 | 489 | 95,51 | 95,51 | **97,96** | 95,51 |
| 9 | 20 | 0,00 | 80,00 | 80,00 | **100,00** |
| 10 | 968 | 68,18 | 80,79 | 80.37 | **86,57** |
| 11 | 2468 | 82,01 | 82,25 | 86,47 | **88,09** |
| 12 | 614 | 64,82 | 81,11 | 85.34 | **87,30** |
| 13 | 212 | 94,17 | **98,06** | 98,06 | 98,06 |
| 14 | 1294 | 92,74 | 95,36 | 93,04 | **95,83** |
| 15 | 380 | 47,59 | 53,01 | 57,83 | **63,25** |
| 16 | 95 | 71,74 | 91,30 | **93,48** | 91,30 |

As mentioned earlier, we apply our proposed method on the AVIRIS 92AV3C where the two-thirds of this image are covered by agricultural land and the one-third by forest or other built structures. According to Table III, it's seen that:

- The homogeneity and energy features disclose the various types classes in the agricultural land compared to the correlation and the contrast that give less efficient results See maps in the following figure7.

- The homogeneity has the best classification accuracy especially in the classes number 9, 13 and 14 as illustrated in figure 7 Image (D).

- The energy on the other hand, gives best results mainly for classes 6, 8 and 16, see figure7 Image (C).

- It's also seen from the results that the correlation gives the maximum classification accuracy of 98.06% for the class 13, figure7 Image (B).

- The class 15 is the weakly classified with 63.25%.

To summarize this step, concerning the separability of classes, we can say that the energy and the homogeneity offer the best potential to distinguish performance of classes in this type of agricultural data even if the number of training pixels is fewer as we see in the above table4 where the classification accuracy achieves 100% for the class 9.

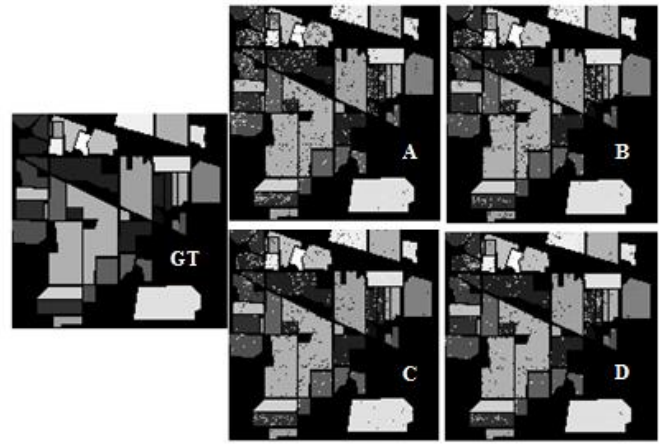

Fig. 7. Original Ground Truth map (GT) and the maps produced by our proposed process3 using Contrast (A), Correlation 5B), Energy (C) and Homogeneity (D).

## V. CONCLUSION

The high dimensionality of the Hyperspectral data imposed many challenging problems in treatment, for this, the dimensionality reduction plies an important role before the classification. Several works were developed in this area but the problematic is always open. In this paper we proposed a new strategy filter combining spectral and spatial information to reduce the dimensionality of HSI and improve the classification results. The GLCM was retained to extract texture features used in our proposed algorithm namely: Contrast, Correlation, Energy and Homogeneity.

We applied our proposed algorithm on the Indiana Pin dataset AVIRIS93AV3C using SVM classifier. The experimental results show the effectiveness selection of the use of both spectral and spatial features with Mutual Information.

This method is very interesting to be investigated and improved considering its performance.